\documentclass[sigconf,nonacm]{acmart}
\AtBeginDocument{%
  }

\usepackage{multirow}
\usepackage{listings}
\usepackage{xcolor}
\usepackage{float}
\usepackage{upquote}
\usepackage{romannum}
\definecolor{CodeBG}{HTML}{F7F8FA}
\definecolor{CodeFrame}{HTML}{E5E7EB}
\definecolor{CodeNum}{HTML}{9CA3AF}
\definecolor{CodeKw}{HTML}{7C3AED}
\definecolor{CodeStr}{HTML}{0369A1}
\definecolor{CodeCmt}{HTML}{6B7280}
\definecolor{CodeOp}{HTML}{0EA5E9}
\definecolor{CodeDef}{HTML}{0F766E}

\lstdefinestyle{prompt}{
  language=Python,
  basicstyle=\scriptsize\ttfamily\linespread{1.05}\selectfont,
  keywordstyle=\color{CodeKw}\bfseries,
  commentstyle=\color{CodeCmt}\itshape,
  stringstyle=\color{CodeStr},
  identifierstyle=\color{black},
  backgroundcolor=\color{CodeBG},
  frame=single,
  framerule=0.6pt,
  rulecolor=\color{CodeFrame},
  framesep=6pt,
  breaklines=true,
  breakatwhitespace=true,
  tabsize=2,
  showstringspaces=false,
  keepspaces=true,
  columns=fullflexible,
  upquote=true,
  captionpos=b,
  aboveskip=6pt,
  belowskip=6pt,
  moredelim=**[is][\color{CodeDef}\bfseries]{§}{§}, 
  emph={self,True,False,None},
  emphstyle=\color{CodeOp}\bfseries
}

\begin{document}

\title{Using Vision-Language Models as Proxies for Social Intelligence in Human-Robot Interaction}

\author{Fanjun Bu}
\email{fb266@cornell.edu}
\orcid{0000-0002-9953-7347}
\affiliation{%
  \institution{Cornell University, Cornell Tech}
  \city{New York}
  \state{NY}
  \country{USA}
}

\author{Melina Tsai}
\orcid{0009-0006-6939-9613}
\affiliation{%
  \institution{Independent Contributor}
  \city{New York}
  \state{NY}
  \country{USA}
}

\author{Audrey Tjokro}
\orcid{0009-0008-0820-2957}
\affiliation{%
  \institution{Cornell University, Cornell Tech}
  \city{New York}
  \state{NY}
  \country{USA}
}

\author{Tapomayukh Bhattacharjee}
\affiliation{%
  \institution{Cornell University}
  \city{New York}
  \state{NY}
  \country{USA}
}

\author{Jorge Ortiz}
\affiliation{%
  \institution{Rutgers University}
  \city{New Brunswick}
  \state{NJ}
  \country{USA}
}
\author{Wendy Ju}
\email{wendyju@cornell.edu}
\orcid{0000-0002-3119-611X}
\affiliation{%
  \institution{Cornell Tech, \\ Jacobs Technion Cornell Institute}
  \city{New York}
  \state{NY}
  \country{USA}
}

\renewcommand{\shortauthors}{Bu et al.}

\begin{abstract}
Robots operating in everyday environments must often decide when and whether to engage with people, yet such decisions often hinge on subtle nonverbal cues that unfold over time and are difficult to model explicitly. Drawing on a five-day Wizard-of-Oz deployment of a mobile service robot in a university café, we analyze how people signal interaction readiness through nonverbal behaviors and how expert wizards use these cues to guide engagement. Motivated by these observations, we propose a two-stage pipeline in which lightweight perceptual detectors (gaze shifts and proxemics) are used to selectively trigger heavier video-based vision-language model (VLM) queries at socially meaningful moments. We evaluate this pipeline on replayed field interactions and compare two prompting strategies. Our findings suggest that selectively using VLMs as proxies for social reasoning enables socially responsive robot behavior, allowing robots to act appropriately by attending to the cues people naturally provide in real-world interactions.
\end{abstract}

\keywords{Human–robot interaction, vision–language models, nonverbal social cues, interaction timing, field deployment, proxemics}

\begin{teaserfigure}
  \includegraphics[width=\textwidth]{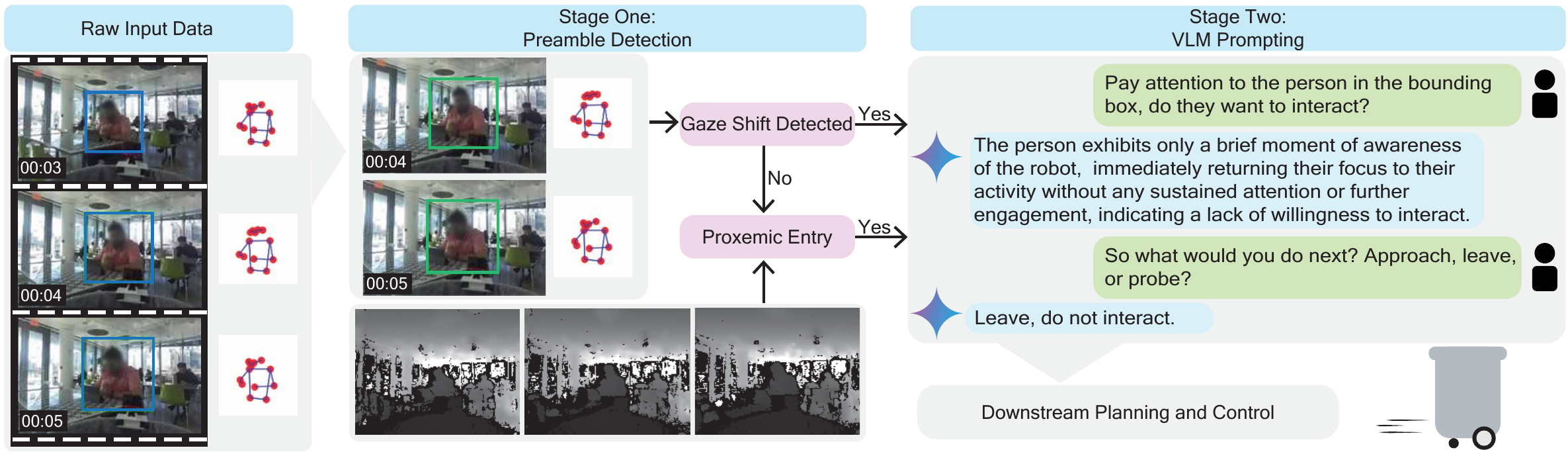}
    \caption{Overview of the two-stage pipeline. Stage \Romannum{1} detects preambles---brief gaze shifts or proxemic entries---that mark socially relevant moments and identify individuals who may be signaling the robot. These triggers initiate Stage \Romannum{2}, where a video-based VLM reasons about the observed scene and produces an evidence-grounded behavior log that supports action selection.}
  \Description{The teaser figure shows our two-stage pipeline as a flow diagram. On the left, Stage \Romannum{1} highlights examples of preambles: a person turning their gaze toward the robot (shown with a bounding box) and a proxemic transition where the robot enters a person’s personal space (illustrated with a depth map). Arrows connect these detections to Stage \Romannum{2} on the right, where a Vision–Language Model takes in the video frame and generates text describing the social scene. The example text log notes that a person briefly looks at the robot and then ignores it, which the system interprets as a signal to leave. The figure emphasizes how low-level visual cues are translated into high-level, interpretable decisions about whether the robot should engage, disengage, or probe.}
  \label{fig:teaser}
\end{teaserfigure}

\maketitle

\section{Introduction}
How does a service robot in a crowded space know which people it should approach? People in similar service situations usually look for non-verbal social responses to their proximity---the turning of the body, or avoidance of gaze---which signal whether service is welcome. These cues are typically implicit and peripheral, making them well-suited for everyday environments where an excess of explicit signaling would be overwhelming \cite{ju2022design}. We use the term \textit{social intelligence} to describe the human ability to "read the room" and make decisions based on subtle, context-dependent cues.

Developing this form of social intelligence in robots has remained a longstanding challenge in HRI. Unlike utilitarian task models with clear reward structures, social interaction is highly contextual and often ambiguous. The same gesture may invite engagement, signal discomfort, or go unnoticed depending on timing, environment, or cultural convention. Designing machine learning systems that account for this complexity through hand-crafted rules or fully enumerated data appears intractable.

Recent advances in video-based vision–language models (VLMs) suggest an alternative pathway. These models demonstrate emergent competence in identifying gestures, gaze shifts, and interpersonal dynamics from naturalistic video, drawing on broad cultural priors embedded in foundation model training. Rather than requiring task-specific data collection and labeling, VLMs offer the possibility of zero-shot social interpretation---robots that can adjust their behavior based on observed cues rather than rigid scripts.

In this work, we explore whether and how it is possible to achieve social intelligence in human–robot interaction by selectively leveraging video-based VLMs. Using a naturalistic Wizard-of-Oz field study of a mobile service robot in a café, we analyze how people respond to the robot’s presence and how expert wizards use those responses to decide when and how to engage. Motivated by these observations, we propose a two-stage framework in which (1) simple temporal features identify socially relevant moments, and (2) a video-based VLM interprets the resulting clip and guides robot action. Our goal is not to solve social intelligence outright, but to examine how VLMs can function as proxies for social reasoning, enabling more fluid, negotiated, and context-aware interaction in real-world settings.
\section{Related Work}
This work addresses a longstanding research topic in HRI---how best to build socially appropriate robots to operate in and amongst people---using recent developments in AI technology.
Previous HRI research has shown that it is possible, within controlled experimental settings, to improve human-robot interaction by adapting proxemic behaviors based on the body signals from the human partner~\cite{mitsunaga2005robot, millan2023proxemic}. We were intrigued by the demonstrated zero-shot ability of large language models and vision–language models (LLMs, VLMs) to interpret social signals~\cite{lee2025human}. Perhaps these tools could allow us to perform social adaptation without the data collection and reinforcement learning required in prior work.

\subsection{Using Foundation Models to Achieve Robot Autonomy}
Foundation models trained on internet-scale data show strong generalization to novel, complex environments, making them compelling alternatives to task-specific learning in robotics \cite{kawaharazuka2024real, firoozi2025foundation}. Many subfields of robotics, including sensing, planning, and control, are leveraging foundation models to achieve full autonomy~\cite{jang2024unlocking}. For instance, \citet{zhang2025nava} uses a VLM to ground natural-language instructions in 3D scenes for navigation, while \citet{gao2024physically} fine-tunes a VLM with physical concepts to improve grasping. Vision Language Action (VLA) systems can output robot actions directly from vision and language—effectively unifying perception, planning, and control \cite{kim2024openvla, black2024pi_0, team2025gemini}. 

 VLMs have also demonstrated potential capability in interpreting social interactions \cite{lee2025human, chakraborty2025vibe}. \citet{williams2024scarecrows} propose “LLMs as Scarecrows,” positioning LLMs as replacements for human wizards in Wizard-of-Oz (WoZ) studies where people are conversing with a robot, and early evidence suggests that LLM-powered robots can achieve social intelligence comparable to wizarded baselines in brainstorming tasks \cite{vrins2024wizard}. Other work explores whether foundation models can infer interaction readiness from images \cite{sasabuchi2025agreeing} or generate socially normative robot actions directly from natural-language descriptions of the scene or behaviors \cite{GenerativeExpressiveRobot}.  

While prior studies demonstrate that VLMs and LLMs can interpret social cues or support conversational HRI, these works predominantly operate on static images, text interactions, or curated datasets. They do not examine situated, temporally unfolding nonverbal cues arising in real-world encounters, nor do they investigate how and when such models should be invoked within an embodied control pipeline. Overall, existing systems typically apply foundation models when explicitly prompted by humans, without mechanisms for selective triggering based on preambles that signal social relevance. Our work addresses these gaps by (1) grounding VLM reasoning in temporally anchored social cues identified from in-the-wild WoZ data, and (2) evaluating VLM-driven action selection on real café encounters, thereby operationalizing foundation models within an interactive, embodied setting.

\subsection{Social Cues and Proxemics in HRI}
The current work also represents novel lines of thinking in HRI about how to equip robots with socially intelligent behavior---in this case, appropriate proxemic behavior.

Early HRI work on proxemics in human-robot interactions focused on determining typical proxemic zones, and identifying individual, contextual, and cultural factors which influence proxemic responses~\cite{takayama2009influences, walters2009empirical, mumm2011human, joosse2014lost, koay2014social}. These types of studies tended to find the proxemic zone threshold through empirical experimentation in controlled laboratory or "living lab" spaces. 

Contemporary researchers, however, have noted the limits of that rationalist approach~\cite{brinck2016making,lawrence2025role}. As ~\citet{brinck2016making} note, "But this gets things backwards–social norms emerge from behavior and depend on social expectations, not the opposite."  Hence, more recent research has focused less on understanding all the situational parameters that influence how a robot should behave, and more on helping the robot to interactively adapt its behavior in response to social cues and signals (such as facial response, gaze, and head movements, body pose changes, and hand gestures) from humans in-situ~\cite{mitsunaga2005robot, stiber2023using, lyu2025signaling, neggers2022determining}.

The interactive approach still requires an understanding of the interaction structure, which might have some context specificity. For example, \citet{brown2024trash} proposes \textit{"spontaneous simple sequential systematics"} to characterize how people interact with trash-barrel robots through motion-based cues and signaling in public plazas. Brown argues that, taken together, the interactions people spontaneously had with the robot had a simple, sequential, and systematic nature. They note that there is a preceding set of conditions, which they call "pre-"---for example, a person gazing at the robot in the square---followed by physical actions and gestures---like waving food wrappers --- that have meanings---offering trash---for the joint action between the people and the robot. This framing provides a foundation for discretizing interactions and motivates a computational pipeline composed of a pattern recognizer followed by a commonsense reasoner.

\section{Empirical Data Collection}
\label{sec:data_collection}
Inspired by prior work using trash barrel robots to elicit naturalistic human-robot interactions in public settings \cite{yang2015experiences, bu2023trash}, we deployed a robot with similar form factors in a semi-public campus café for five days. Through Wizard-of-Oz deployment, we observe unscripted interactions between the robot and unsuspecting participants.

\begin{figure}[bht]
\centering
\includegraphics[width=\linewidth]{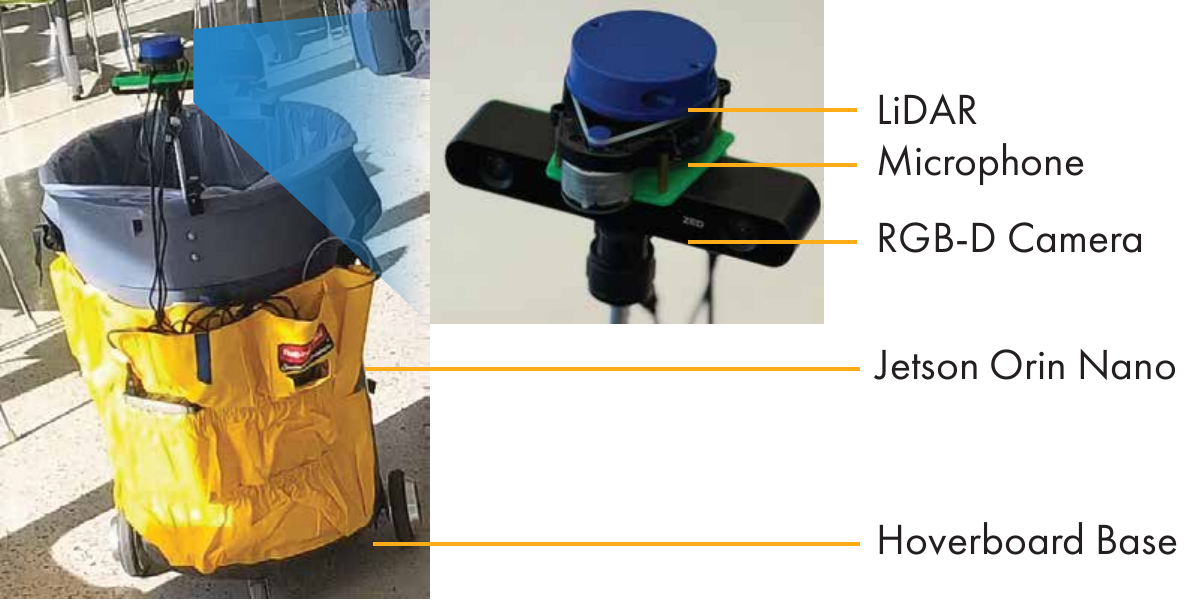}
\caption{Robot Hardware Setup.}
\Description{The figure illustrates the physical setup of our trash-barrel–shaped mobile service robot used in the study. The robot’s body resembles a cylindrical barrel mounted on a hoverboard-based mobile chassis. Around the trash barrel is a yellow caddy bag, within which sits a Jetson Orin Nano computer that serves as the main onboard processor. A 2D LiDAR sensor, a forward-facing RGB-D camera, and a microphone are mounted above the barrel facing forward.}
\label{fig:sensor_suite}
\end{figure}

\subsection{Robot Platform}
Our robot builds upon Bu et al.’s trash barrel robot design \footnote{Found online at \url{https://github.com/IRL-CT/Summer25_Trashbot}}, with a 32-gallon BRUTE barrel mounted on a hoverboard base \cite{bu2023trash}.
To support richer perception and interaction, we added a ZED-2i RGB-D camera, a ReSpeaker microphone array, and a 2D LiDAR. For computation, we replaced the Raspberry Pi 4 with an NVIDIA Jetson Orin Nano, housed with its power supply in a caddy bag due to its larger size and power demands (Fig.~\ref{fig:sensor_suite}). The system runs Ubuntu 22.04 with ROS 2 Humble, enabling 15 fps image streaming and real-time control. 

\subsection{Protocol}
We conducted our study in a university café frequented by students and local residents. Deployments occurred daily around 13:00 during peak lunch traffic, with each session lasting one to two hours depending on activity levels.

The robot was controlled via a Wizard-of-Oz protocol: a researcher teleoperated it in real time using a joystick, guided exclusively by the robot’s onboard audio-visual feeds. During the deployment, the researcher sat inconspicuously in a corner of the café, appearing to be a student working on a laptop.

This study was approved by the university’s Institutional Review Board (IRB). Flyers were posted at all entrances and throughout the café around the recording area, informing participants that research activities and video recording were taking place. By entering the space, participants voluntarily consented to the terms and conditions of the study. Participants were also provided with the research team’s contact information on the flyers and informed of their right to request removal of their data at any time.

Unlike the protocol used by Bu et al.~\cite{bu2023trash, pelikan2025people}, where the operator responded naturally to social context, our protocol was designed to elicit a broader range of social signals, including edge cases. The operator systematically navigated the robot in continuous loops through the café, deliberately probing by approaching individuals and groups regardless of whether they appeared engaged, occupied, or socially unavailable. This included intrusive behaviors, such as approaching from behind or interrupting conversations. In effect, the setup created a stage for negotiation, with the robot approaching in the user’s general direction to signal availability for interaction.

By relaxing the usual constraint of socially appropriate WoZ control, we aimed to capture the full spectrum of human responses to robotic presence, from receptive to avoidant. This approach supports models that generalize across real-world scenarios, including ambiguous or non-norm-compliant moments. Interactions remained bounded by strict safety constraints: the robot moved slowly, made no physical contact, and any impropriety was limited to the social domain. Note that not every approach was intentional— sometimes participants were simply along the robot’s path—but the underlying principle of probing intent remained intact.

\subsection{Dataset}
\label{sec::dataset}
Across the five-day field deployment, the robot initiated 106 approach attempts and interacted with 174 participants, corresponding to approximately 150 unique individuals. Interaction durations ranged from 2 to 31 seconds, with a mean length of 13.3 seconds. Of these encounters, 43 involved a single person and 62 involved small groups. In terms of outcomes, 42 interactions concluded with participants disposing trash into the robot, while 64 ended with the robot disengaging, where people showed no interest in interaction.

All data were recorded as ROS bags containing synchronized RGB images, depth images, microphone audio, and the wizard’s control commands (Twists). These bags provide a temporally aligned multimodal record of each interaction episode.

We divided the dataset into training and testing partitions based on entire days to avoid cross-day leakage and maintain natural variation. The training set consists of 76 interactions drawn from Days two to four, while the test set contains 30 interactions from Days one and five. This partitioning preserves temporal coherence while ensuring that the test set contains data from distinct days with different traffic patterns and participant populations.

\subsection{Expert Annotations}
\label{sec:annotation}
Two expert researchers with extensive experience in Wizard-of-Oz studies annotated recorded first-person video data showing the robot probing participants. Both researchers have more than 20 hours of experience with wizarding robots in the wild. Combining bottom-up observations from the deployment and ~\citet{brown2024trash}'s formulation of spontaneous, simple, sequential systematics, we segment interactions into preamble (what Brown et al. termed "pre-"), cue (physical means of interacting), and signal (meaning of the cue under context).

Formally, we define \textit{preamble}, \textit{cue}, and \textit{signal} as the following:
\begin{itemize}
\item \textit{Preamble}: action that sets preconditions/onset for information exchange. 
\item \textit{Cue}: overt actions that convey intention, such as gaze aversion to signal disinterest or a wave to invite engagement.
\item \textit{Signal}: the interpretation of a cue within its situational and cultural context, aligning with the definition of signals in \citet{fiore2013toward}.
\end{itemize}

\begin{table}[h!]
\centering
\caption{Examples for Preamble, Cue, and Signal}
\begin{tabular}{@{}lll@{}}
\toprule
\textbf{Preamble} & \textbf{Cue} & \textbf{Signal} \\ \midrule
\multirow{1}{*}{\begin{tabular}[c]{@{}l@{}}Gaze-shift \\ Proxemic entry\end{tabular}} 
  & \begin{tabular}[c]{@{}l@{}}Wave \\ Body Posture Change \\ Look Away \\ Continue prior activity\end{tabular}
  & \begin{tabular}[c]{@{}l@{}}Request service \\ Reject service \\ Actively ignoring \\ Unaware of robot \end{tabular} \\
\bottomrule
\end{tabular}
\label{tab:social-cues}
\end{table}

Annotations were made at fixed two-second intervals. At each interval, annotators labeled whether a preamble or a cue was present in the clip, and when a cue was present, they also recorded its meaning as a signal. Many cues can convey the same semantic meaning (e.g., both waving and nodding can mean welcoming the robot). In addition, annotators judged whether, based on the participant’s nonverbal behavior, the robot can make a decision whether to approach at that moment (decision timing), and labeled each clip as keep probing (undecided), approach, or leave. By design, all clips preceding the one containing a cue were marked as "keep probing."

We adopted a two-second interval labeling scheme rather than frame-by-frame annotation to better accommodate the subtle and sometimes temporally diffuse nature of non-verbal social cues. Behaviors such as active ignoring often unfold gradually over time. The fixed interval approach helps mitigate subjectivity in pinpointing such transitions and promotes consistency across annotations.

Our annotation protocol reveals that ``Undecided'' corresponds to a robot action (status quo, probing to elicit more evidence) rather than a hidden human state. In deployment scenarios with ambiguous social cues, the optimal response is often temporal deferral.

 The initial labeling of decision timing between annotators yielded a Cohen’s kappa of 0.81, indicating strong agreement beyond chance. We focus on agreement for decision timing because the choice of robot action is contingent on when a decision is deemed possible. In cases of disagreement, we resolved conflicts by taking the earlier decision timing. This reflects a design choice: we prefer to bias the robot toward acting slightly earlier rather than being overly conservative and delaying its response.

By inspecting the annotation, we found that the average distance at which gaze-shift preambles occurred follows a roughly normal distribution, peaking at about 1.2 meters (4 feet) (see Appendix \ref{appendix:proxemics}). Notably, this aligns with Edward Hall’s proxemics theory, marking the boundary between the personal and social zones\cite{hall1966hidden}.

\section{Two-Stage Pipeline}

We formalize the robot's decision-making problem as a two-stage gated inference pipeline that operates on discrete temporal segments. This framework accurately reflects our implementation, which uses discriminative vision-language models for per-clip inference without recursive belief propagation or policy optimization. See Figure~\ref{fig:teaser} for a visual overview of the pipeline.

\subsection{Problem Formulation}

The robot operates in discrete time intervals indexed by $t \in \{1, 2, \ldots, T\}$, where each interval corresponds to a fixed 2-second video clip. At each timestep $t$, the robot observes visual data $o_t$ (RGB-D frames with pose keypoints) and must select an action $a_t \in \mathcal{A}$ from a discrete action space.

We model the person's latent intent as a binary state $s_t \in \mathcal{S} = \{\text{Interact}, \text{DoNot}\}$. Within each 2-second clip, we assume the state $s_t$ remains static, which reflects our annotation protocol and the temporal granularity of social cue interpretation. Decisions are made independently per clip based solely on the current observation $o_t$, with no information carried forward from previous clips. This per-clip independence means the robot does not maintain a recursive belief state $b_t$ that accumulates evidence across timesteps.

\subsection{Two-Stage Gated Decision Process}

Our pipeline consists of a binary gate function $g: \mathcal{O} \to \{0, 1\}$ (Stage \Romannum{1}) and a conditional inference function $f: \mathcal{O} \to \mathcal{A}$ (Stage \Romannum{2}) that operates only when the gate triggers.

\begin{figure}[h!]
    \centering
    \includegraphics[width=\linewidth, trim={0 0 0 0.7cm},clip]{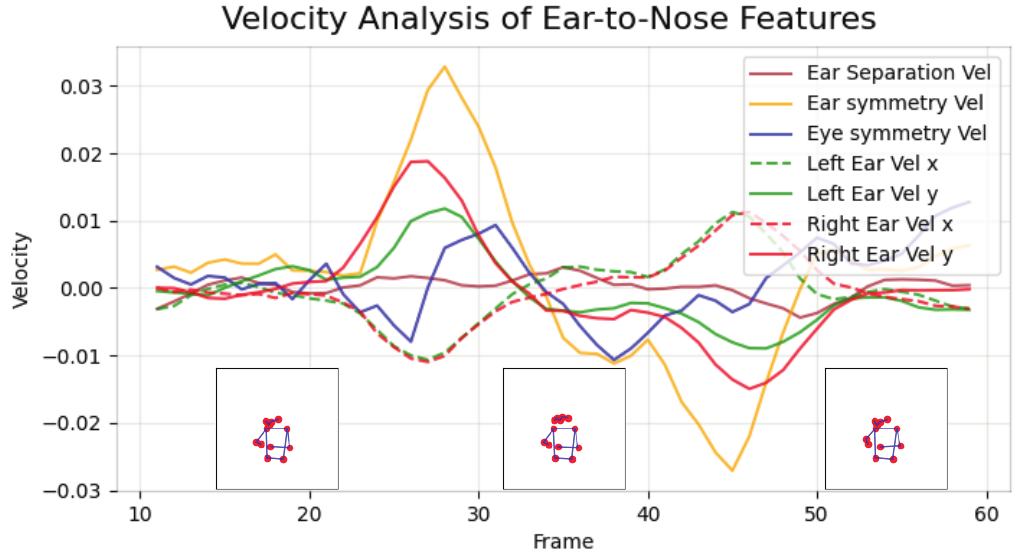}
    \caption{Velocity signals during a brief glance shift toward and away from the robot (same interaction as Fig.~\ref{fig:teaser}).}
    \label{fig:velocity_signal}
    \Description{The figure plots the velocity of ear-to-nose and eye symmetry features over time. The x-axis represents frame number, and the y-axis represents velocity in normalized pixel space. Two green lines show the velocity of the left ear (solid line: y-coordinate, dashed line: x-coordinate). Two red lines show the velocity of the right ear (solid line: y-coordinate, dashed line: x-coordinate). 
    Three solid lines show the velocities of the ear separation (brown), ear symmetry (yellow), and eye symmetry (blue). The plot fluctuates around zero with both positive and negative values. Two clear peaks appear around frame numbers 35 and 45, corresponding to gaze-shift moments. Three upper-body keypoint maps are overlaid on the figure, showing the person’s posture before the gaze shift, during the shift toward the camera, and after turning away from the camera.}
\end{figure}

\paragraph{Stage \Romannum{1}: Preamble Detection Gate}
Stage \Romannum{1} implements a rule-based detector that identifies preambles (proxemic transitions and gaze shifts) that signal potential social engagement. Formally, we define the gate function as:
\begin{equation}
\label{eq:gate}
g(o_t) = \begin{cases}
1 & \text{if } \text{preamble\_detected}(o_t) \\
0 & \text{otherwise}
\end{cases}
\end{equation}
where $\text{preamble\_detected}(o_t)$ returns true when either:
\begin{itemize}
\item A gaze shift is detected via a trained histogram-based gradient boosting classifier operating on head pose velocity features (see Figure~\ref{fig:velocity_signal} for example velocity signals), or
\item A proxemic entry occurs (robot enters personal zone $\leq 1.2$m) \textit{only} when no prior gaze-based preamble detection has occurred within the same 2-second window.
\end{itemize}

The proxemics trigger fires exclusively when the robot enters the personal zone without a preceding gaze shift, capturing cases where participants either did not notice the robot or deliberately continued their activity to signal disinterest.

When $g(o_t) = 0$, the robot defaults to action $a_t = \text{Probe}$ without invoking the computationally expensive VLM. This design reduces VLM calls from 487 (if querying every detected person every 2 seconds) to 129 in our deployment (see Section~\ref{sec:eval-stage-one}), while preserving decision quality by focusing inference on moments with likely social relevance.

\paragraph{Stage \Romannum{2}: Conditional VLM Inference}
\label{Conditional_VLM_Inference}
When $g(o_t) = 1$, Stage \Romannum{2} invokes a vision-language model to analyze the clip and produce an action. The VLM operates as a discriminative model, directly estimating the posterior distribution over intent:
\begin{equation}
\label{eq:discriminative}
q_t(s) \approx P(s \mid o_t)
\end{equation}
where $q_t(s)$ is the VLM's output distribution over $\mathcal{S}$ given observation $o_t$. Note that this is not a generative observation model $Z(o_t \mid s)$ as required for POMDP belief updates. Instead, the VLM supplies discriminative posterior estimates directly, without converting them into likelihoods. This aligns with our approach of prompting the VLM to output an intent estimate together with the supporting observable evidence. Specifically, the VLM generates two outputs via the prompts detailed in Appendix~\ref{appendix:prompts}:
\begin{enumerate}
\item A behavior log $\ell_t$ with timestamped evidence of observable social cues, formatted as $[\text{mm:ss}]:[\text{evidence}]$, produced by the independent analysis prompt (Listing~\ref{lst:prompt-individual-gaze} or \ref{lst:prompt-individual-proxemics}).
\item An intent estimate $\hat{s}_t \in \mathcal{S}$ inferred from the behavior log through self-consistency or self-critique synthesis (Listing~\ref{lst:conrtadiction_prompt} and ~\ref{lst:self_critique_prompt}, or ~\ref{lst:prompt-consistency}).
\end{enumerate}

These outputs are then passed to a another VLM prompt (Listing~\ref{lst:prompt-action}) that generates the final action $a_t = f(o_t, \ell_t, \hat{s}_t) \in \mathcal{A}$. This prompt-based action generation replaces the role of a POMDP policy $\pi^*(b_t)$ that maps belief states to actions via utility maximization.

\subsection{Uncertainty Quantification}

Social cues can be ambiguous, leading to inconsistent VLM outputs. We address this through two uncertainty quantification strategies that operate on multiple independent VLM samples.

\paragraph{Self-Consistency Uncertainty}
In the self-consistency approach, we sample $K=5$ independent behavior logs $\{\ell_t^{(1)}, \ldots, \ell_t^{(K)}\}$ from the VLM with temperature $\tau = 0.7$ using the prompts in Appendix~\ref{appendix:prompts}. For each log, we extract intent predictions $\{\hat{s}_t^{(1)}, \ldots, \hat{s}_t^{(K)}\}$.
We define epistemic uncertainty as the disagreement across samples:
\[
u_{\text{SC}} = 1 - \max_{s \in \mathcal{S}} \frac{n_s}{K}
\]
where $n_s$ is the number of samples predicting state $s$. When $u_{\text{SC}} > \eta$ for threshold $\eta$, the system defaults to $\text{Probe}$ due to high uncertainty. This uncertainty measure is \textit{not} a calibrated probability suitable for POMDP policy optimization; it serves as a heuristic for deferral, as evaluated in Section~\ref{sec:eval-stage-two}.

The final behavior log is synthesized via majority voting (Listing~\ref{lst:prompt-consistency}): behaviors mentioned in $\geq 4/5$ samples are included, with timing determined by the most frequently mentioned timestamps.

\paragraph{Self-Critique Uncertainty}
In the self-critique approach, the same $K=5$ independent samples are generated, then analyzed for contradictions using the contradiction identification prompt (Listing~\ref{lst:conrtadiction_prompt}). The VLM identifies disputed claims $\mathcal{D}$ among the samples, then verifies each claim against the video evidence using the verification prompt (Listing~\ref{lst:self_critique_prompt}).
Uncertainty arises from:
\begin{itemize}
\item Contradictions that remain after video verification
\item ``Inconclusive'' judgments caused by limited or ambiguous evidence
\end{itemize}

If contradictions remain unresolved or key claims are inconclusive, the system again defaults to $\text{Probe}$. Like self-consistency uncertainty, this is a heuristic deferral mechanism rather than a calibrated probability distribution over states.

Both uncertainty measures capture epistemic uncertainty (disagreement about interpretation) rather than aleatoric uncertainty (inherent ambiguity in the social signal). Critically, these measures are \textit{not} calibrated probabilities suitable for decision-theoretic policy optimization; they serve as heuristics for when to defer decisions.

\subsection{Action Selection}

Action selection occurs via prompt-based generation from the VLM (see Listing~\ref{lst:prompt-action} in Appendix~\ref{appendix:prompts}). Given the synthesized behavior log $\ell_t$ and intent estimate $\hat{s}_t$, the VLM is prompted to output an action $a_t \in \{\text{Approach}, \text{Leave}, \text{Probe}\}$ with a brief justification.

This process does \textit{not} involve solving a cost-optimal policy $\pi^*(b_t)$ over belief states (as would be required in a POMDP). Instead, action selection relies on the VLM's implicit social reasoning, guided by the behavior log and intent estimate. The $\text{Probe}$ action serves as both the default (when $g(o_t) = 0$) and the fallback (when uncertainty is high or evidence is ambiguous), mirroring the temporal deferral strategy used by human annotators (Section~\ref{sec:annotation}).

\section{Operationalizing the Two-Stage Framework}
To translate the two-stage framework into a working system, we implement Stage \Romannum{1} as a lightweight perceptual detector of social preambles and Stage \Romannum{2} as a VLM-based interpreter of social context.

\subsection{Stage \Romannum{1} Implementation}
Following our problem formulation, we design a rule-based system that handles each preamble type independently.

\paragraph{Data Preprocessing}
Given the limited size of our dataset, we applied extensive preprocessing to maximize the utility of each clip. Individuals were tracked with YOLO-11x, and 2D body keypoints were extracted using ViTPose++ \cite{yolov11, xu2023vitpose++}. Pose data were normalized by re-centering keypoints at the shoulder midpoint and scaling them by the inverse torso length, defined as the Euclidean distance between the shoulder and hip midpoints. Distances to users were estimated by averaging depth values from a neighborhood around their shoulder midpoints.

\paragraph{Proxemic Thresholding}
In our pipeline, individuals whose distance from the robot remains beyond four meters throughout the 2-second window are excluded, as interaction is unlikely. This threshold reflects our observation that only one participant in the dataset exhibited a gaze shift at four meters, whereas all others did so within three meters (Appendix \ref{appendix:proxemics}). Conversely, when the robot enters an individual’s personal zone (within 1.2 meters) without eliciting a gaze-based signal, we invoke the VLM to reason about likely scenarios such as active ignoring. This conditional invocation balances efficiency with interpretability in uncertain social contexts.

\paragraph{Gaze Shift Detection}
 We train a histogram-based gradient boosting classifier to detect gaze-shift preambles. Training data consists of 2-second clips labeled by expert annotators as either containing a clear gaze-initiated preamble (positive class) or being a resting state (negative class), as discussed in Section \ref{sec:annotation}. To balance the training data, negative examples are sampled from the 2-second segments immediately preceding each annotated gaze-shift preamble. We split our five-day dataset into training and testing partitions to prevent data leakage, as discussed in \ref{sec::dataset}. Within the training portion, we conducted a five-fold cross-validation to perform a grid search over hyperparameters, optimizing for the F1 score.

\begin{table}[h!]
\renewcommand{\arraystretch}{1.5} 
\centering
\caption{Features used for the preamble detection model.}
\begin{tabular}{ll}
\toprule
\textbf{Feature} & \textbf{Formula} \\ \midrule
    Left Ear to Nose (x, y) & \(\overrightarrow{\mathbf{P}_{\text{Left Ear}} - \mathbf{P}_{\text{Nose}}}\) \\
    Right Ear to Nose (x, y) & \(\overrightarrow{\mathbf{P}_{\text{Right Ear}} - \mathbf{P}_{\text{Nose}}}\) \\
    Eye Separation & \(\|\mathbf{P}_{\text{Left Eye}}- \mathbf{P}_{\text{Right Eye}}\|\) \\
    Ear Separation & \(\|\mathbf{P}_{\text{Left Ear}} - \mathbf{P}_{\text{Right Ear}}\|\) \\
    Ear Symmetry & \(\frac{min(\|\overrightarrow{\mathbf{P}_{\text{Left Ear to Nose}}}\|, \|\overrightarrow{\mathbf{P}_{\text{Right Ear to Nose}}}\|)}{max(\|\overrightarrow{\mathbf{P}_{\text{Left Ear to Nose}}}\|, \|\overrightarrow{\mathbf{P}_{\text{Right Ear to Nose}}}\|)}\) \\
\bottomrule
\end{tabular}
\label{tab:stage-one-features}
\end{table}

Because gaze shifts are brief and subtle, we focus on head-related velocity signals. 
The features whose velocities are computed are shown in Table \ref{tab:stage-one-features}.

\(\mathbf{P}_{\text{keypoint}}\) represents the 2D coordinate of the keypoint in pixel space after normalization. Each feature is of length $T - 1$ (where \(T = 15 \text{fps} * 2s = 30\)), corresponding to a two-second observation window. We compute the temporal derivatives (velocities) of each feature and extract summary statistics: maximum, minimum, and standard deviation over time. This results in a 21-dimensional feature vector serving as input to the boosting classifier.

Because our keypoints are in 2D, we primarily use ear-based features to approximate head rotation, as they provide more salient cues than eye-based features. However, ear keypoints are more susceptible to jitter from occlusions, which can produce abrupt, spurious changes. To address this, we also include eye separation as an auxiliary feature. Although highly correlated with ear separation, eye separation is generally less noisy and provides a smoother signal. Incorporating both improves robustness, allowing the model to better distinguish true head-turns from detection artifacts and enhancing classification performance.

\subsection{Stage \Romannum{2} Implementation}
\label{sec:stageTwo}
In Stage \Romannum{2}, candidate clips identified in Stage \Romannum{1} are analyzed with a VLM to infer participants’ intentions. To spatially ground the reasoning, the focal individual is highlighted with a bounding box---blue if selected by a gaze shift, orange if selected by a proxemic entry. Each gaze-shift segment is padded by two seconds before and after the detected preamble to capture temporal context, while each proxemic segment includes the two seconds following the robot’s entry into the personal zone. Triggers occurring within one second are grouped as a single event. We also use trigger-specific prompts for gaze-shift and proxemic detections (see Appendix~\ref{appendix:prompts}), providing the VLM more information about its interaction contexts.

To capture temporal dynamics, the bounding box turns green whenever the boosting algorithm detects a potential gaze shift. The VLM is explicitly prompted to focus on the highlighted individual and frames where the bounding box turns green. This approach is similar to set-of-marks prompting, which has been shown to improve the alignment of textual outputs with specified visual referents \cite{yang2023set}. Because we require a single decision per participant, each video prompt highlights only one individual.
\section{Evaluation}
To evaluate the two-stage pipeline under realistic conditions, we re-ran the system on the recorded ROS bags from the test-set interactions, effectively replaying the original scenarios. For each interaction, we manually segmented a continuous clip that captured the entire episode: starting from the robot’s initial probe toward a given interactant and ending when the interaction was naturally resolved. These replay clips averaged 11.7 seconds in duration (range: 5.2–22.2 seconds).

We then executed the full pipeline end-to-end on each replay, allowing Stage \Romannum{1} to trigger VLM queries based on detected social cues and allowing Stage \Romannum{2} to produce scene interpretations. A domain expert inspected the resulting logs, intent estimates, and chosen actions to qualitatively assess whether the system’s interpretations aligned with the observed human behavior. Because VLM queries rely on external APIs, our evaluation emphasizes semantic correctness--whether the system made the right decision---rather than real-time performance.

\subsection{Stage \Romannum{1} Preamble Detection Performance}
\label{sec:eval-stage-one}
The best performing boosting model was evaluated once on the held-out test set for gaze-based preamble detection. The best-performing hyperparameters are: learning rate = 0.1, maximum depth = 7, number of iterations = 100, and minimum samples per leaf = 8. This configuration achieved an F1 score of 0.79 on the validation set and 0.80 on the test set, with a test accuracy of 0.77 and a test ROC AUC of 0.86. The confusion matrix is $\begin{pmatrix}
15 & 6\\ 
4 & 20
\end{pmatrix}$.

To contextualize these results, we evaluated the Stage \Romannum{1} detector on 30 unseen interaction episodes containing 53 ground-truth interactants. The boosting model identified 71 candidate interactants, of which 47 were true interactants. At the episode level, Stage \Romannum{1} correctly selected the main interactant in 26 of the 30 clips. With respect to timing, the detector triggered at the correct preamble moment for 43 interactants, indicating that the velocity-based features were generally effective at capturing the onset of social relevance.

Across all episodes, Stage \Romannum{1} produced 112 gaze-triggered calls and 17 proxemic-triggered calls. This imbalance reflects the pipeline design: when the robot enters the personal zone, people usually respond with a gaze shift. Therefore, proxemic transitions mainly captured cases where participants either did not notice the robot or deliberately continued their activity to signal disinterest.

\begin{figure}[h]
    \centering
    \includegraphics[width=\linewidth]{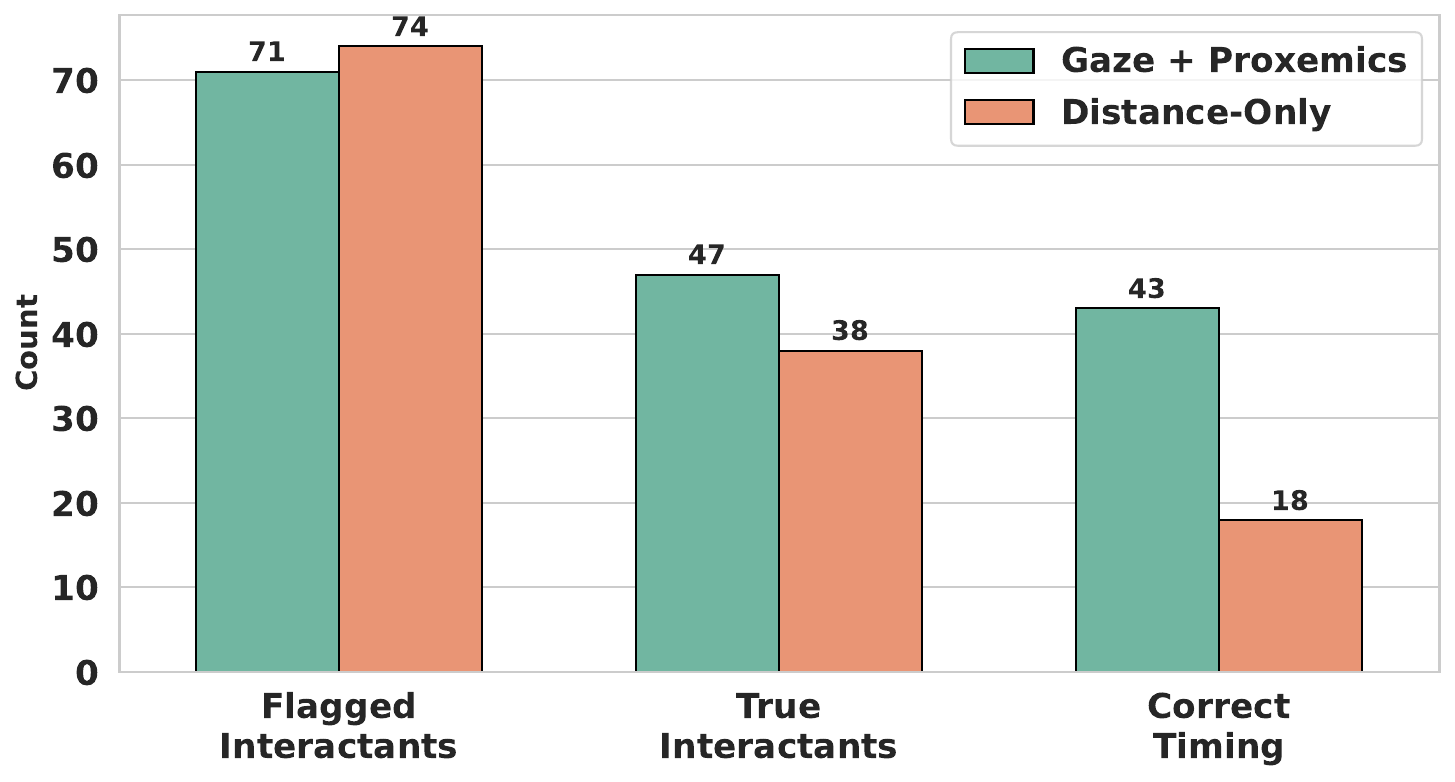}
    \caption{Comparing Stage \Romannum{1} performance with a distance-only baseline.}
    \label{fig:baseline}
    \Description{Bar chart comparing our Stage \Romannum{1} performance with a distance-based heuristic. See results in text.}
\end{figure}

We compare these results to a distance-only baseline that prompts the VLM using a 2-second clip captured immediately after the robot enters a person’s personal zone (1.2 m; Fig. \ref{fig:baseline}). Under this heuristic, the system flagged 74 interactants, of which 38 were true interactants. At the episode level, 27 of the 30 main interactants were selected; however, the heuristic triggered at the correct preamble moment for only 18 interactants, substantially lower than the 43 correct timings achieved by Stage \Romannum{1}. One the other hand, prompting the VLM every two seconds for every detected person in the scene would have resulted in 487 calls on the same dataset. Thus, Stage \Romannum{1} reduces unnecessary VLM invocations by an order of magnitude while improving temporal precision relative to the distance-only baseline.

\subsection{Stage \Romannum{2} VLM Performance}
\label{sec:eval-stage-two}
In evaluating the second stage, we assess both the generated behavior logs and the final action outputs. An expert wizard, who also participates in the initial data annotation discussed in section \ref{sec:data_collection}, reviewed each VLM output, checking whether the descriptions matched the observed behaviors and whether the resulting user intent and robot actions aligned with human judgment.

\begin{figure*}[htbp]
    \centering
    \includegraphics[width=\linewidth]{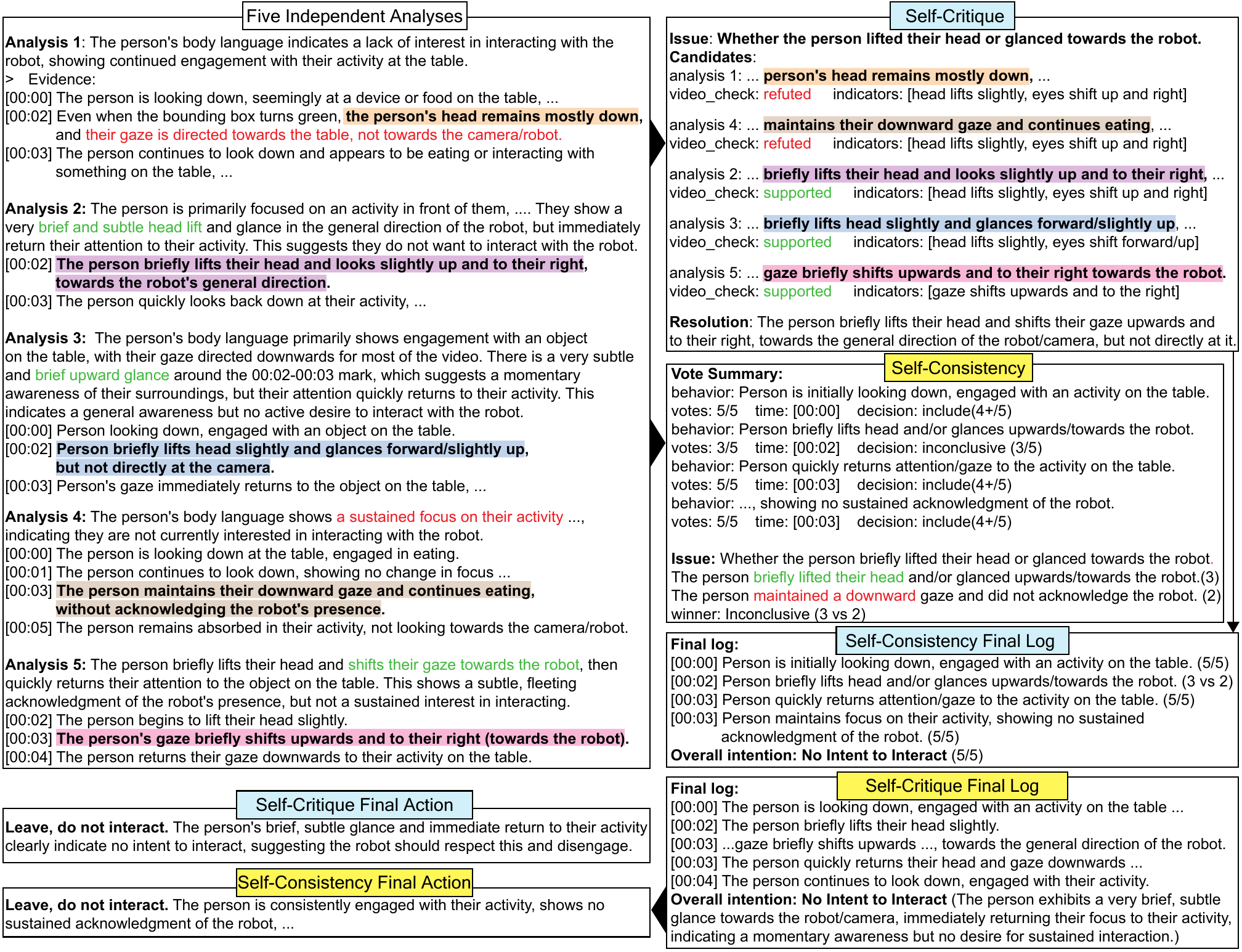}
    \caption{Excerpted VLM outputs from different reasoning strategies for interaction (Figure \ref{fig:teaser}).}
    \Description{Image is text-based and self-explanatory.}
    \label{fig:prompting}
\end{figure*}

For proxemics-based calls, accuracy is relatively high, as these inputs typically lack subtle social cues by design. Out of 17 proxemics-based calls, both self-consistency and self-critique approaches correctly handled 14 (82\%).
For gaze-based calls, both prompting strategies produced richer scene descriptions than a single-prompt baseline (Fig.~\ref{fig:prompting}). After excluding four low-quality clips, we evaluated 108 gaze-based calls. Semantic accuracy on behavior logs was 60\% (65/108) for both approaches, although each approach was correct on different subsets of clips. Notably, in the self-consistency approach, the model exhibits high uncertainty on more clips than in the self-critique approach (22 vs 11).

Despite modest semantic accuracy, both approaches achieve good action-level performance. Self-critique reached an accuracy of 73\% (79/108), and self-consistency has an accuracy of 74\% (80/108). Probing behavior is considered correct when the scene is ambiguous, and false when a clear signal exists. However, some rare cases allow for multiple valid interpretations—for example, a person fixing their gaze on the robot may signal either welcome or curiosity, making both “approach” and “probe” plausible. Among the probing actions caused by high uncertainty, self-critique generated 11, of which 5 were judged accurate, whereas self-consistency generated 22, of which 10 were judged appropriate. We observed that, despite hallucinations and missed cues, the model’s uncertainty generally does not affect the overall signaling from participants. This is directly observable in the self-consistency approach, where the outputs can have high variance on individual behaviors, such as a brief glance, but the analyses on a person's intention can be unanimous (Fig. \ref{fig:prompting}). 

\section{Discussion}
\subsection{An Alternative Path to Social Generalization}
Although our two-stage pipeline is intentionally simple, it illustrates a broader design principle for social robots: VLMs can serve as proxies for social reasoning, allowing robots to act appropriately without constructing explicit, domain-specific models of human intent. Rather than requiring exhaustive training data or hand-engineered rules, the robot leverages the embedded social priors in a pretrained VLM—summoned only at moments identified as socially meaningful by lightweight perceptual cues.

The classic story of Clever Hans provides a useful reminder that interaction is shaped not only by internal reasoning but also by sensitivity to subtle social cues. Hans, a horse once believed to perform arithmetic, was in fact responding to nonverbal signals from nearby humans. Although often cited as a cautionary tale about misleading correlations \cite{lapuschkin2019unmasking}, the episode also underscores how much information people communicate through small, often unconscious movements. Our use of this example does not imply accidental success; rather, it highlights the value of designing robots that can attend to the same kinds of social cues that people rely on in everyday coordination.

Viewed through this lens, our framework’s probe action operationalizes social negotiation. Rather than assuming the robot can infer intent from a single observation, probing allows the system to gather additional evidence (cues), test hypotheses about a person’s receptivity, and respond to unfolding signals. The robot does not---and perhaps cannot---maintain a fully specified model of human values or preferences. Yet it can still act appropriately by attending to how people gesture, orient, or disengage in the moment.

\subsection{Using VLMs to Analyze Social Signals} 
In this work, we investigated how to use VLMs as a proxy to social intelligence in social human–robot interaction. As shown in the results, our pipeline can reproduce decision-making patterns similar to those used by human wizards in the wild. While only 60\% of generated behavior logs matched human annotations, the action-level outcomes are promising: the robot can plausibly rely on the generated actions to make social interaction decisions.

Although Stage \Romannum{1} detection was tailored to our trash-disposal setting, Stage \Romannum{2} was connected to the domain only through the text prompt and consistently produced outputs that captured the task space. This design blueprint, which combines specialized cue detection with general-purpose reasoning, provides a practical template for injecting social intelligence across application domains. We acknowledge limitations in testing only one robot form factor and location, and anticipate that adapting Stage \Romannum{1} will be necessary when audiences, roles, or designs elicit different responses. However, compared to learning-based methods, our approach leverages zero-shot VLM capabilities, yielding interpretable reasoning, eliminating retraining for new users, and supporting flexible shifts in operating models across diverse contexts.

One of the reasons we can analyze how the VLMs succeed and fail is that language itself is a part of input and output. This makes the process inspectable, which also helps us bring in non-technical experts to help us analyze and adapt the robot to new contexts.

Our approach demonstrates the feasibility of using the general reasoning power of VLMs to approximate wizard-level social intelligence. Future work will explore fine-tuning and distilling large VLMs into smaller, domain-focused reasoning systems that run efficiently on the edge. More broadly, our results suggest that framing interaction as negotiation offers a robust paradigm: it captures the dynamic nature of interaction, integrates uncertainty, and situates behaviors within a unified framework for analysis and discussion.

\subsection{Methodological and Design Considerations}
This paper presents a proof of concept for the overall approach of using VLMs to interpret social signals, but several design choices deserve mention. Some were guided by pragmatic cost considerations. For instance, we sampled five responses from the VLM for each prompt, though in principle, more samples could yield greater robustness. The use of bounding-box color to highlight temporal saliency and guide VLM generation was another pragmatic choice, tested experimentally and found to be effective. We also selected Gemini-Flash-2.5 for its strong performance in video understanding, while a systematic comparison of different VLMs remains beyond the scope of this work.

Our pipeline combines specialized detection with generalized reasoning. For our application, false positives are tolerable because the reasoning stage can filter them out, but noisy detection systems still involve a trade-off between accuracy and computation. A higher false-positive rate results in more prompts to the VLM and thus greater computational cost. At the same time, we observed that users naturally repeat and amplify their signals if the robot fails to respond. For example, a person may offer a second dismissive gaze if the robot continues to approach after missing the first. This redundancy provides a practical margin of error for false negatives.

\paragraph{The Value of Social Preambles}
One might expect the first stage of our system to be a full action-recognition module capable of detecting socially meaningful gestures such as waving, pointing, or beckoning. However, social cues in natural environments are often brief, small in magnitude, and highly context dependent. In our early prototypes, we experimented with state-of-the-art action detection models, but these models proved brittle when applied to low-resolution, crowded café footage and could not reliably detect the fleeting, subtle movements characteristic of social preambles. Curating a sufficiently large and representative dataset for fine-tuning such models was also not feasible in our setting.

Our findings highlights the utility of preambles---brief, early nonverbal cues such as gaze shifts and proxemic entries---as stable anchors for social interpretation. Preambles generalize better and can be detected robustly with lightweight, velocity-based features. Although this approach is intentionally minimal, it is precisely this simplicity that makes Stage \Romannum{1} effective in real-world field conditions: it avoids overfitting to fine-grained gestures and instead focuses on reliable triggers that indicate when deeper, VLM-based social reasoning should be invoked.

\paragraph{Designing Useful VLM prompts}
It is also worth noting that the second stage of the pipeline is highly sensitive to prompt design.  We found several strategies that improve both performance and interpretability. For independent analyses, we encouraged diversity in reasoning to capture a broad range of possible interpretations before converging on a final answer. Explicitly asking the model to cite evidence with timestamps improved accuracy, as timestamps help the self-critique or self-verification process by locating contradictions in time. 
Conversely, imposing excessive constraints or forcing the model to output its most confident answer reduced performance, as fleeting but important social cues could be overlooked.

\paragraph{Self-consistency vs. self-critique}
A key difference between our self-consistency and self-critique approaches lies in how they handle uncertainty. Self-consistency exposes disagreement among the five independent analyses directly as an uncertainty signal, whereas self-critique resolves contradictions through verification and only surfaces ambiguity when the evidence itself is inconclusive. This makes self-critique more decisive, since disagreements are collapsed during verification. However, this decisiveness does not necessarily translate into higher accuracy.

The choice between exposing variance or collapsing it is ultimately a design decision, and each strategy has complementary strengths. Self-critique is more prone to hallucinations, with five invented behaviors carried into outputs compared to one for self-consistency, but it can capture fleeting cues that self-consistency would miss. Self-consistency is more robust against hallucinations yet cannot recover when subtle cues are overlooked by most analyses. It also defaults to probing more often (22 vs. 11) since high disagreement among its five analyses triggers deferral. In contrast, self-critique resolves variance through verification, shifting uncertainty to the semantic interpretation of ambiguous behaviors, which occurs less often.

\subsection{Contextual Insights for Social HRI}
While we believe this approach can generalize to applications in other human-robot interaction domains, we will note some specific aspects of this application that influence its performance. First of all, trash collection is a very subjective matter; the seemingly simple question of what counts as trash depends on the owner’s perspective~\cite{strasser2000waste}.   It is not possible to passively determine whether a person has trash; hence, applying the VLMs to interpret social signals makes more sense than interpreting the physical evidence at hand. For this reason, the robot must negotiate by signaling availability and interpreting responses to determine whether people have trash to collect.

The subjectivity of the trash collection task poses challenges for annotation, as different wizarding experts exhibit distinct operating styles that introduce inconsistencies in labeling. Most discrepancies arise when it is unclear whether the person noticed the robot—for example, when it approached from behind or when cues were deliberately withheld. Some wizards refrained from approaching, while others continued until no space remained. This inconsistency highlights the inherent difficulty of grounding robotic perception in socially contextualized signals. On the other hand, the ambiguity can be resolved by continued probing. Also, people change their decisions in real time. Hence, framing the interaction as ongoing negotiation rather than one-shot decision-making enables adaptation to ambiguity and shifting situations.
\subsection{Limitations}
\label{sec:limitations}
While we designed our analysis pipeline so that it could be applied in settings beyond our trash collection setting, we acknowledge the limitation that we have not yet tried the pipeline for other tasks or settings. Our aim for this work was to establish a proof-of-concept for automatic recognition of social signals in a realistic field setting. 

Our evaluation reports semantic and action accuracy but does not establish whether the VLM outputs provide calibrated probabilities suitable for policy use. These omissions mean that the effect size of the VLM cannot yet be quantified in absolute decision-theoretic terms. Future work can address calibration (e.g., through temperature scaling, expected calibration error, or Brier scores) and introduce simple heuristic baselines to place the results on a firmer comparative footing.

\section{Conclusions}
This work demonstrates a design approach for endowing robots with social intelligence by leveraging VLM-based reasoning at key moments shaped by people’s in situ responses. Integrating lightweight social cue detection with temporally grounded VLM interpretation enables a form of intelligence that is pragmatic and adaptive—one that treats interaction as a process of continual negotiation rather than a fixed decision rule.

More broadly, our findings show how nonverbal cues can structure human–robot interaction and how they can be systematically incorporated into robot behavior without requiring explicit models of human preferences or norms. We offer this framework as a step toward social intelligence in open-world contexts, where what counts as appropriate behavior must be discovered through interaction. We hope this work contributes to the development of robots that act not only more competently, but also more contextually and socially attuned in everyday environments.

\begin{acks}
This work is sponsored by NSF \#2423127 FRR: Social and Contextual Models of Interaction with Everyday Robots.
\end{acks}

\bibliographystyle{ACM-Reference-Format}
\bibliography{bu}

\appendix
\section{Caveats and Limitations: POMDP Soundness}
\label{sec:caveats-pomdp}

While our problem domain of partially observable decision-making under uncertainty naturally suggests a POMDP formulation, our implementation diverges from standard POMDP assumptions in several critical ways. We document these gaps explicitly to clarify what would be required for full POMDP soundness.

\paragraph{Observation Model}
\textbf{Implemented:} We use VLMs as discriminative models providing posterior estimates $P(s \mid o)$ directly. The VLM outputs text logs and intent classifications (via prompts in Appendix~\ref{appendix:prompts}) without any conversion to generative likelihoods $Z(o \mid s)$. The discriminative posterior $q_t(s) \approx P(s \mid o_t)$ from Equation~\eqref{eq:discriminative} is used directly for per-clip inference, with no conversion to the likelihood form required for POMDP belief updates.

\textbf{For POMDP soundness:} A proper POMDP requires a generative observation model $Z(o_t \mid s_t)$ that specifies the probability of observing $o_t$ given state $s_t$. This would enable Bayesian belief updates via:
\[
b_t(s) \propto Z(o_t \mid s) \sum_{s'} T(s \mid s', a_{t-1}) b_{t-1}(s')
\]
To achieve this, one would need to either: (1) learn a generative model $Z(o \mid s)$ from data, or (2) convert discriminative VLM outputs to likelihoods via a Bayes bridge (e.g., $Z(o \mid s) \propto P(s \mid o) / \pi(s)$ for prior $\pi(s)$), though this requires careful calibration as noted in Section~\ref{sec:limitations}.

\paragraph{Belief Propagation}
\textbf{Implemented:} Decisions are made independently per 2-second clip, as described in Section~\ref{sec:annotation}. There is no recursive belief state $b_t$ that accumulates evidence across timesteps. Each clip is analyzed in isolation, with the VLM producing a fresh posterior estimate $q_t(s) \approx P(s \mid o_t)$ for each clip without incorporating information from previous clips. This matches the evaluation protocol in Section~\ref{sec:eval-stage-two}, where each clip receives an independent action decision.

\textbf{For POMDP soundness:} A proper POMDP maintains a belief state $b_t(s) = P(s_t \mid \mathcal{H}_t)$ where $\mathcal{H}_t = \{o_1, a_1, \ldots, o_t\}$ is the interaction history. Belief updates follow the recursive Bayesian filter $b_t(s) \propto Z(o_t \mid s) \sum_{s'} T(s \mid s', a_{t-1}) b_{t-1}(s')$, allowing the robot to accumulate evidence over time and make decisions that account for temporal dependencies. Our implementation does not maintain such a belief state; instead, each clip's decision depends only on $o_t$.

\paragraph{Transition Model}
\textbf{Implemented:} We assume static state within each 2-second clip, consistent with our annotation protocol (Section~\ref{sec:annotation}). There is no explicit transition model $T(s_{t+1} \mid s_t, a_t)$ specifying how intent evolves over time or in response to robot actions. The static assumption reflects the temporal granularity of social cue interpretation: within a 2-second window, we treat the person's intent as fixed, and decisions are made without modeling how actions might influence future intent states.

\textbf{For POMDP soundness:} A transition model $T(s' \mid s, a)$ is required to predict how the person's intent changes over time and in response to robot actions. This would enable belief propagation via $b_t(s) \propto \sum_{s'} T(s \mid s', a_{t-1}) b_{t-1}(s')$. Such a model could be learned from interaction data or specified based on domain knowledge (e.g., intent may become more negative after repeated ignored approaches). Our implementation does not use transitions; each clip's state is inferred independently.

\paragraph{Policy Optimization}
\textbf{Implemented:} Actions are selected via VLM prompting (Listing~\ref{lst:prompt-action}) based on behavior logs and intent estimates. There is no solved cost-optimal policy $\pi^*(b_t)$, no explicit reward function $R(s, a)$, and no decision thresholds $\tau_*$ derived from utility maximization. Instead, the VLM's action prompt implicitly encodes social reasoning without explicit cost/reward parameters. This matches the evaluation in Section~\ref{sec:eval-stage-two}, where actions are assessed for semantic accuracy rather than utility maximization.

\textbf{For POMDP soundness:} A proper POMDP policy $\pi^*(b_t)$ maximizes expected cumulative reward:
\[
\pi^*(b_t) = \arg\max_{a \in \mathcal{A}} \sum_s b_t(s) \left[ R(s, a) + \gamma \sum_{s', o'} T(s' \mid s, a) Z(o' \mid s') V^*(b') \right]
\]
This requires: (1) specifying reward/cost parameters ($r_{\text{hit}}$, $c_{\text{intrude}}$, $c_{\text{miss}}$, $c_{\text{probe}}$), (2) solving the Bellman equation (via value iteration, policy iteration, or approximate methods), and (3) deriving decision thresholds $\tau_* = c_{\text{intrude}} / (r_{\text{hit}} + c_{\text{miss}} + c_{\text{intrude}})$ that map belief states to actions. Our implementation uses prompt-based action generation instead of this optimization.

\paragraph{Value of Information}
\textbf{Implemented:} Stage \Romannum{1} uses preamble detection (gaze shifts and proxemic transitions) as a heuristic proxy for when VLM inference is worthwhile. There is no explicit computation of expected value of information $\mathbb{E}[\text{VoI}]$ comparing decision quality before and after VLM inference. The gate function $g(o_t)$ in Equation~\eqref{eq:gate} triggers based on detected preambles rather than comparing expected utilities $\max_a U_a(b^{(0)})$ vs $\max_a U_a(b^{(1)})$.

\textbf{For POMDP soundness:} A principled VoI gate would compute:
\[
\text{VoI} = \max_{a \in \mathcal{A}} \sum_s b^{(1)}(s) R(s, a) - \max_{a \in \mathcal{A}} \sum_s b^{(0)}(s) R(s, a)
\]
where $b^{(0)}$ is the prior belief and $b^{(1)}$ is the posterior after VLM inference. The gate would trigger when $\mathbb{E}[\text{VoI}] > c_{\text{vlm}}$, requiring explicit reward modeling $R(s, a)$ and belief comparison. Our implementation uses preamble detection as a proxy, motivated by the intuition that preambles signal moments when additional information is likely to change the decision from the default Probe action.

\paragraph{Probability Calibration}
\textbf{Implemented:} VLM outputs $q_t(s) \approx P(s \mid o_t)$ are used directly without calibration, as noted in Section~\ref{sec:limitations}. The discriminative posterior estimates from Equation~\eqref{eq:discriminative} are not validated for calibration (e.g., via expected calibration error or Brier scores), and the uncertainty measures $u_{\text{SC}}$ and self-critique uncertainty serve as heuristics rather than calibrated probabilities suitable for policy optimization.

\textbf{For POMDP soundness:} Policy optimization requires well-calibrated probability estimates. If $q_t(s)$ systematically over- or under-confident, the resulting policy $\pi^*(b_t)$ will be suboptimal. Calibration methods (temperature scaling, Platt scaling, isotonic regression) can map raw VLM outputs to calibrated probabilities, but this requires validation data and careful evaluation, as explicitly noted in Section~\ref{sec:limitations} as future work.
\section{Appendix: Proxemics}
Our in-field data shows that the distance at which individuals exhibit non-verbal gaze-shift preambles toward the robot approximates a normal distribution, which peaks around 1.2 meters (4 feet), a boundary consistent with Edward Hall’s definition of personal zones.
\label{appendix:proxemics}
\begin{figure}[ht]
    \centering
    \includegraphics[width=\linewidth]{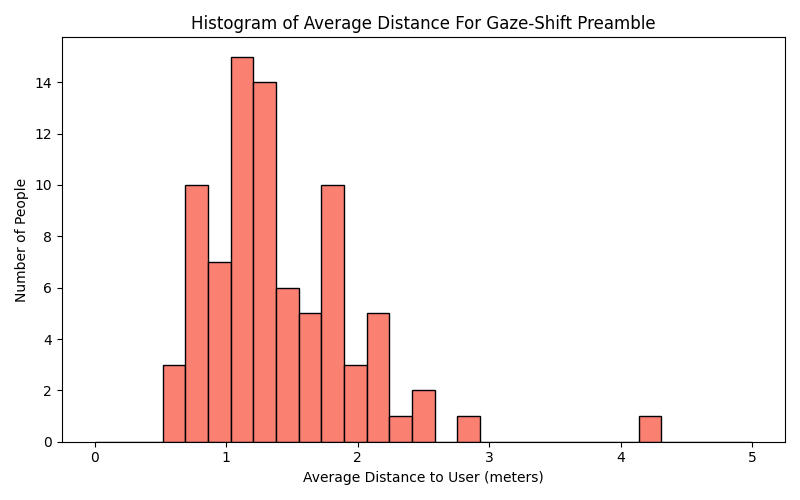}
    \caption{Distance between the robot and the interactant when gaze-shift preambles are observed.}
    \Description{The figure is a vertical bar histogram titled “Histogram of Average Distance For Gaze-Shift Preamble." The horizontal axis (x-axis) represents the average distance from the robot to the interactant, measured in meters, ranging from 0 to 5. The vertical axis (y-axis) represents the number of people, ranging from 0 to 15. The distribution of distances is concentrated between 0.8 and 2.5 meters, forming a bell-shaped, Gaussian-like curve. The highest frequency (the peak) occurs around 1.2 meters, where about 15 people are observed. Beyond 3 meters, very few people exhibited gaze-shift preambles. There is one clear outlier above 4 meters. Overall, the figure shows that gaze-shift preambles most commonly occur when the robot is close, about 1.2 meters (4 feet) from the user.}
    \label{Fig:proxemics}
\end{figure}

\section{Appendix: Prompts used in Stage \Romannum{2}}
\label{appendix:prompts}
In this section, we present all prompts used with Google Gemini-Flash-2.5 during stage two, covering both the self-critique and self-consistency approaches.

\subsection{Independent Analysis Prompt}
Below, we provide the prompts used to generate five independent analyses for each input video. These prompts instruct the VLM to focus on the person within the bounding box and attend closely to moments when potential gaze-shifts are detected.
\begin{lstlisting}[style=prompt,caption={Prompt for five independent analyses (Gaze).},label={lst:prompt-individual-gaze}]
prompt = (
    "You are a robot whose job is to approach people so they can throw away their trash. You are approaching the person shown in the video in the blue bounding box. The video features the person's subtle non-verbal behaviors towards your approach. The camera is co-located with the robot; 'toward the robot' == 'toward the camera'. The bounding box may flicker or disappear, but you should focus on the person in the bounding box throughout the video."
    "Task: Reason step by step. Try to capture all behaviors of the person of interest. Focus on the person's subtle body language. Does the person want to interact with you?"
    "If there are concrete actions that you can pinpoint, you should cite them as evidence. Make sure the evidence is directly visible in the video."
    "Pay special attention when the bounding box turns green, which may suggest a potential cue of interest."
    "Note that the robot is also moving/turning, so the person's relative position may change. Make sure to distinguish between the person's movement and the robot's movement."
    "output format (use exactly):"
    "Answer: [The person's body language..., showing ...]"
    "Evidence: [mm:ss] <brief micro-cues>"
)
\end{lstlisting}

\begin{lstlisting}[style=prompt,caption={Prompt for five independent analyses (Proxemics).},label={lst:prompt-individual-proxemics}]
proxemics_prompt = (
    "You are a robot whose job is to approach people so they can throw away their trash. You are approaching the person shown in the video in the orange bounding box. You are currently in that person's personal zone. The camera is co-located with the robot; 'toward the robot' == 'toward the camera'."
    "The bounding box may flicker or disappear, but you should focus on the person in the bounding box throughout the video."
    "Task: Reason step by step. Try to capture all behaviors of the person of interest. Focus on the person's subtle body language. Does the person want to interact with you?"
    "If there are concrete actions that you can pinpoint, you should cite them as evidence. Make sure the evidence is directly visible in the video."
    "Pay special attention when the bounding box turns green, which may suggest a potential cue of interest."
    "Note that the robot is also moving/turning, so the person's relative position may change. Make sure to distinguish between the person's movement and the robot's movement."
    "output format (use exactly):"
    "Answer: [The person's body language..., showing ...]"
    "Evidence: [mm:ss] <brief micro-cues>"
)
\end{lstlisting}

\subsection{Self-critique and self-consistency prompt}
For the self-critique approach, the VLM was first prompted to identify contradictions among the five independent analyses.
\begin{lstlisting}[style=prompt,caption={Self-Critique. Prompt used to identify contradictions among the five independent analyses.},label={lst:conrtadiction_prompt}]
conrtadiction_prompt = (
    "You are given 5 independent analyses of a video showing a blue/orange-box person reacting to a robot. Your job is to extract all contradictions, disagreements, or disputed claims among the analyses."
    "Do NOT attempt to resolve or fact-check them."
    "For each contradiction, list (format exactly, no JSON is necessary):"
    " - issue: <what is disputed>"
    "   candidates:"
    "   - analysis: <n>"
    "     quote: \"<verbatim>\""
    "   ... (repeat for each analysis with a distinct claim)"
    "Only include issues where there is a clear disagreement or mutually exclusive claim."
    "If all analyses agree, output: contradictions: []"
    "Here are the 5 analyses:"
)
\end{lstlisting}

We then re-prompted the VLM with a verification prompt, instructing it to check the identified contradictions against the video and synthesize a final behavior log.
\begin{lstlisting}[style=prompt,caption={Self-Critique. Prompt used to verify contradictions against the video and produce a consolidated behavior log.},label={lst:self_critique_prompt}]
self_critique_prompt = (
    "You are given 5 independent analyses of a video showing the person in the blue/orange bounding box reacting to a robot."
    f"Video duration: {video_duration_sec} seconds. The video is 15 fps."
    "Here are the 5 analyses:"
    + "".join([f"[Analysis {idx+1}/5]:{analysis}" for idx, analysis in enumerate(analyses)])
    + "Here are the contradictions extracted from the analyses:"
    + f"{contradiction_text}"
    + "ROLE: Verifier. Your job is to synthesize a clear, human-interpretable behavior log of the blue-box/orange-box person's nonverbal behaviors and intent, using both the video and contradiction analyses."
    + "Instructions:"
    + "- Focus on the person in the blue/orange bounding box only."
    + "- Do NOT hallucinate. Only include behaviors or intent that are directly visible."
    + "- Do NOT make up new claims outside the existing analysis unless you have solid video evidence to support them."
    + "- Fact-check every contradiction in the provided contradictions against what is visible in the video."
    + "- For each candidate analysis, state whether it is 'supported', 'refuted', or 'inconclusive', and cite specific visual evidence (body part movement, direction, duration, micro-cues, with timestamps)."
    + "- If a candidate analysis is ambiguous or not clearly supported, mark it as 'inconclusive' and explain."
    + "- If any analysis mentions a brief or subtle cue (e.g., a quick glance, fleeting gesture, or micro-expression), carefully check the video for this event, even if only one analysis notices it. If visible, include it in the final log with supporting evidence and a timestamp."
    + "- At the end, state the person's overall intention to interact with the robot, based ONLY on directly observable behaviors."
    + "CHECKLIST FOR EACH CANDIDATE ANALYSIS:"
    + "- Is this claim visible in the video? (Allow a plus or minus 1-second grace window.)"
    + "- If not, mark it as 'inconclusive' or 'refuted' and explain."
    + "- For 'supported' claims, cite the exact visual evidence."
    + "- If conflicting claims are both supported by video evidence, mark the resolution as 'inconclusive'."
    + "OUTPUT FORMAT:"
    + "contradictions:"
    + " - issue: <what is disputed>"
    + "   candidates:"
    + "   - analysis: <n>"
    + "     quote: \"<verbatim>\""
    + "     video_check: supported | refuted | inconclusive"
    + "     indicators: [<brief visual indicators>]"
    + " - resolution to the issue: <one line>. State inconclusive if conflicting claims are supported by the video."
    + "Final log (Verification):"
    + " - [mm:ss] <verified claims> "
    + " - [mm:ss] <inconclusive claims, and explain what is the inconclusive part>"
    + "Overall intention: [Interact | No Intent to Interact | Inconclusive] (rationale: <one line>)"
)       
\end{lstlisting}

For self-consistency, we prompted the model to generate five independent analyses and then applied a majority-vote scheme across them to produce the final behavior log.
\begin{lstlisting}[style=prompt,caption={Self-Consistency prompt},label={lst:prompt-consistency}]
self_consistency_prompt = (
    "You are given 5 independent analyses of a video showing a person reacting to a robot. ROLE: Majority Vote Synthesizer. Your job is to combine the 5 analyses using majority voting."
    "MAJORITY VOTE RULES:"
    "1. For each behavior/claim, count how many analyses support it (out of 5)."
    "2. Include behaviors supported by majority (4+/5)."
    "3. Exclude behaviors supported by minority (1/5) ."
    "4. For contradictory claims, the majority position wins."
    "5. If uncertain (e.g., 3 vs 2, 2 vs 3), mark claim as inconclusive and include the more supported position."
    "PROCESS:"
    "- Go through behaviors/claims mentioned across all 5 analyses."
    "- Count votes for each behavior."
    "- Include only majority-supported behaviors (4+/5)."
    "- For timing, use the most frequently mentioned time range."
    "- For overall intention, count votes for [Interact | No Intent to Interact | Inconclusive]. If uncertain (3 vs 2, 2 vs 2), mark claim as [Inconclusive]."
    "OUTPUT FORMAT:"
    "vote_summary:"
    "  - behavior: <description>"
    "    votes: <n>/5"
    "    time: <most common time mentioned>"
    "    decision: include(4+/5) | exclude(1/5) | inconclusive (2/5 or 3/5)"
    "contradictions:"
    "  - issue: <what is disputed>"
    "    position_A: \"<description>\" (votes: <n>/5)"
    "    position_B: \"<description>\" (votes: <n>/5)"
    "    winner: position_A | position_B | Inconclusive (if 3 vs 2 or 2 vs 2)"
    "Final log (Majority Vote):"
    "  - [mm:ss] <majority behavior> (votes: <n>/5)"
    "  - [mm:ss] <inconclusive claims, and vote summary (e.g. 3 vs 2)>"
    "Overall intention: Overall intention: [Interact | No Intent to Interact | Inconclusive] (votes: <n>/5)"
    "Here are the 5 analyses:"
    + "".join([f"[Analysis {idx+1}/5]:{analysis}" for idx, analysis in enumerate(analyses)])
)
\end{lstlisting}

\subsection{Action Prompt}
Given the behavior logs produced by self-consistency and self-critique, we prompted the VLM once more to generate high-level robot actions (approach, leave, or probe) based on the interpreted social scene.
\begin{lstlisting}[style=prompt,caption={Action Prompt},label={lst:prompt-action}]
action_prompt = (
    "Given the following overall intention of a person's reaction to the robot, output the appropriate robot action towards the person as either 'Approach to interact', 'Leave, do not interact', or 'Inconclusive, Keep probing'."
    "Justify your answer in 1-2 sentences."
    f"{log_text}"
    "Decision:"
)
\end{lstlisting}
\end{document}